%% file: main.tex
\documentclass{article}
\usepackage[main,final]{neurips_2026}

\usepackage[utf8]{inputenc} 
\usepackage[T1]{fontenc}    
\usepackage{hyperref}       
\usepackage{url}            
\usepackage{booktabs}       
\usepackage{amsfonts}       
\usepackage{nicefrac}       
\usepackage{microtype}      
\usepackage{xcolor}         

\usepackage{amsmath, amssymb, amsfonts}
\usepackage{algorithm}
\usepackage{algpseudocode}
\usepackage{multirow}
\usepackage{booktabs}
\usepackage{tikz}
\usetikzlibrary{matrix, fit, backgrounds, shapes.geometric}
\usepackage{pgfplots}
\usepgfplotslibrary{groupplots}
\pgfplotsset{compat=1.18}
\usepackage{subcaption}

\title{CMPQ: Compensatory Quantization via Input-Aware Hessian Damping}

\author{%
  Deng Pan \\
  University of Notre Dame \\
  \texttt{dpan@nd.edu} \\
  \And
  Grigorii Khvatskii \\
  University of Notre Dame \\
  \texttt{gkhvatsk@nd.edu} \\
  \And
  Xiaobao Huang \\
  University of Notre Dame \\
  \texttt{xhuang2@nd.edu} \\
  \And
  Peiyu Li \\
  University of Notre Dame \\
  \texttt{pli9@nd.edu} \\
  \And
  Shangqian Gao \\
  Florida State University \\
  \texttt{sg24bi@fsu.edu} \\
  \And
  Ting Hua \\
  University of Notre Dame \\
  \texttt{thua@nd.edu} \\
  \And
  Nitesh V. Chawla \\
  University of Notre Dame \\
  \texttt{nchawla@nd.edu}
}

\begin{document}

\maketitle

\begin{abstract}
As large language models (LLMs) scale, post-training weight quantization (PTQ) has become a standard tool for deploying them on commodity hardware with little loss in quality. However, existing quantization solvers such as GPTQ calibrate each weight matrix against a corrupted signal: the calibration input observed at any deep layer is the output of the cumulative quantized chain, not a clean full-precision reference.  We propose \textbf{CMPQ (Compensatory Quantization)}, which reframes the corruption as evidence about \emph{which input directions have become unreliable} and tells the solver to stop spending its compensation budget on them. CMPQ measures the divergence between the quantized and full-precision input and uses it to reshape the curvature that GPTQ already relies on, redirecting compensation onto weights attached to clean input subspaces and cutting the propagation path of upstream noise. On the Qwen3 family at three sizes (4B / 8B / 14B), CMPQ improves over GPTQ and is competitive with GPTAQ at 3-bit; at 2-bit, where GPTQ and AWQ collapse to near-random accuracy, it outperforms all baselines. Code is available at \url{https://github.com/pd90506/CMP-Q}.
\end{abstract}

\input{sections/introduction-v2}
\input{sections/related-work}
\input{sections/preliminaries}
\input{sections/methodology}
\input{sections/experiments}

\newpage
\bibliographystyle{unsrt}
\bibliography{reference}

\appendix
\input{sections/appendix-derivation}
\input{sections/appendix-additional}

\end{document}

%% file: sections/introduction-v2.tex
\section{Introduction}
\label{sec:intro}

Modern large language models have gained substantial capabilities by
scaling their parameter count to hundreds of billions. For example, the LLaMA-3 family ranges up to 405B
parameters~\cite{grattafiori2024llama}, Qwen3 spans dense models up to 32B and an MoE flagship of
235B~\cite{yang2025qwen3}, and DeepSeek-R1 totals 671B parameters~\cite{guo2025deepseek}. However, this significant scaling comes with a memory footprint that quickly
saturates the high-bandwidth memory of even server-class hardware,
making both deployment and inference economically inefficient.

Post-training weight quantization (PTQ)~\cite{gholami2022survey} has
emerged as one of the most effective remedies. Among PTQ approaches,
finetuning-free methods compress pre-trained weights to low-bit
representations (4-bit, 3-bit, or even 2-bit) using only a small
calibration set, preserving most downstream quality without
gradient-based retraining. For weight-only finetuning-free PTQ,
\textbf{Hessian-based solvers}~\cite{frantar2022optimal,frantar2022gptq,nagel2020up,li2021brecq,dettmers2023spqr,shao2023omniquant,chee2023quip,tseng2024quip}
are the dominant approach: they use second-order information from the
calibration input to identify weight perturbation directions that matter
most and compensate for quantization error accordingly.
GPTQ~\cite{frantar2022gptq} is the canonical example, optimizing the
weight matrix column by column to minimize quantization error. A
parallel line of work exemplified by AWQ~\cite{lin2024awq} migrates
quantization difficulty via per-channel activation-aware scaling rather
than residual compensation; we include it as an orthogonal baseline.

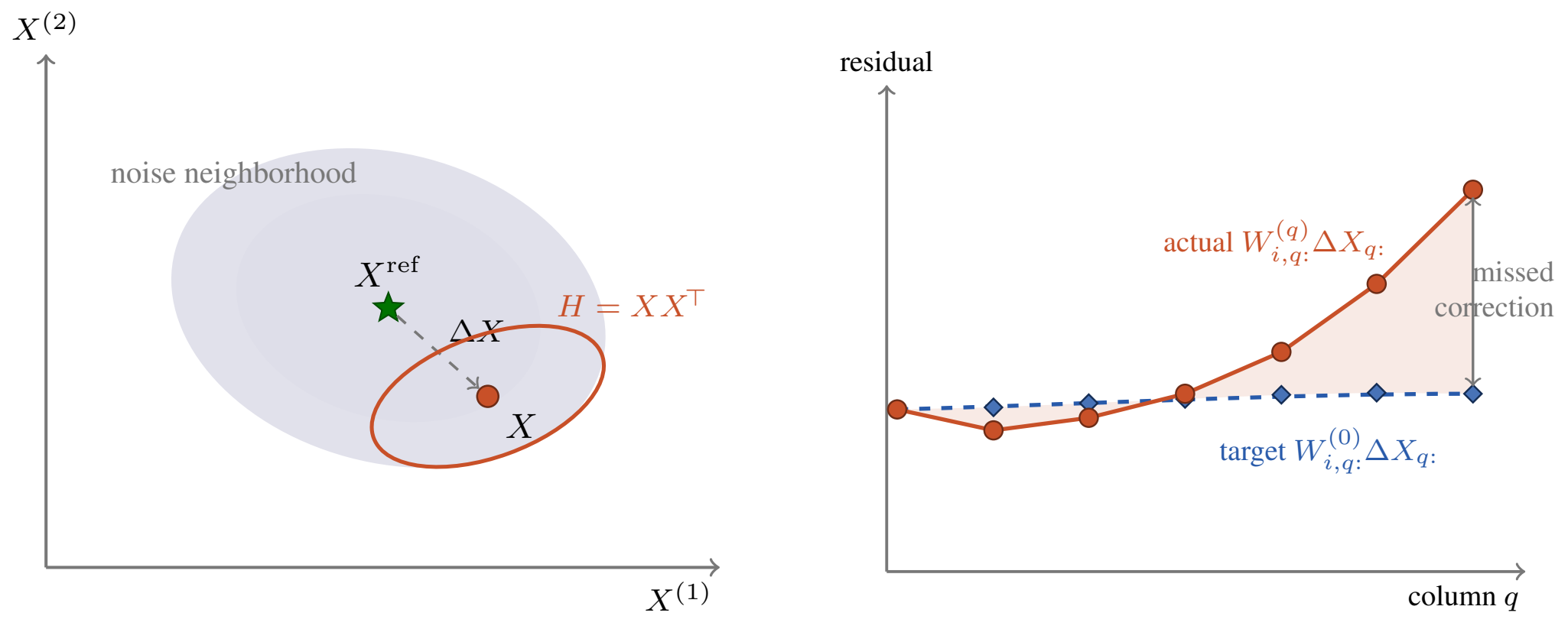
\begin{figure}[t]
  \centering
  \begin{subfigure}[t]{0.48\textwidth}
    \centering
     \makebox[\linewidth][c]{%
      \resizebox{\linewidth}{!}{\input{figures/gptq_blind}}}
    \subcaption{}
    \label{fig:gptq-blind}
  \end{subfigure}\hfill
  \begin{subfigure}[t]{0.48\textwidth}
    \centering
    \makebox[\linewidth][c]{%
      \resizebox{\linewidth}{!}{\input{figures/gptaq_drift-v2}}}
     \subcaption{}
    \label{fig:gptaq-drift}
  \end{subfigure}
  \caption{Two failure modes of existing weight-only solvers under
  upstream noise. \textbf{(a)} GPTQ builds its Hessian from a single
  corrupted realization $X$, anchored at the wrong center and blind to
  the surrounding upstream-noise neighborhood (haze). \textbf{(b)}
  GPTAQ's per-column residual is defined against the running weight
  $W^{(q)}$, but the correct target is the original $W^{(0)}$. For
  $q \ge 1$, $W^{(q)}$ has accumulated error-propagation updates from
  earlier columns, so the residual GPTAQ applies (orange) drifts away
  from the residual it should apply (dashed blue); the missed-correction
  gap (shaded) compounds across columns.}
  \label{fig:fig1}
\end{figure}

At the core of these solvers is a per-layer reconstruction objective: the
quantized weights should reproduce the full-precision layer's output on a
small calibration set. The input observed at calibration time, however,
is the activation produced by the cumulative quantized chain, not the
clean activation the full-precision model would produce. The two differ
by an \emph{upstream-noise} term that varies token by token, so what
calibration sees is a single realization $X$ drawn from a wider noise
neighborhood around the FP16 reference $X^{\mathrm{ref}}$. Existing
solvers silently treat this realization as if it were the clean
reference. GPTQ in particular builds its Hessian from $X$ alone, so the
resulting curvature is anchored on one corrupted observation and remains
blind to the rest of the neighborhood (Fig.~\ref{fig:gptq-blind}).

A natural response is to compare the full-precision layer's output on
the clean input against the quantized layer's output on the corrupted
input. GPTAQ~\cite{li2025gptaq} does exactly this, deriving a
closed-form per-column residual correction. The correction is
theoretically sound, but its closed form is tied to the running weight
$W^{(q)}$ rather than the original $W^{(0)}$. For $q \ge 1$, $W^{(q)}$
has accumulated error-propagation updates from previously quantized
columns and inevitably departs from $W^{(0)}$; the residual GPTAQ
applies thus targets the wrong reference, and the missed correction
compounds across columns, eroding the
solution~\cite{li2026rethinking} (Fig.~\ref{fig:gptaq-drift}).

In this work, we propose \textbf{CMPQ (Compensatory Quantization)},
which sidesteps both failure modes by acting on the curvature itself.
Rather than fitting to a single realization or chasing it with a
per-column residual, CMPQ treats upstream noise as evidence of
\emph{which input directions have become unreliable} and folds this
signal directly into the Hessian that the GPTQ solver already relies on.
Noisy directions then appear with inflated curvature, so the column-wise
solver allocates less compensation to weights attached to them and
concentrates its budget on the more reliable subspace.

We summarize our main contributions as follows:
\begin{itemize}
  \item \textbf{Noise-aware Hessian formulation.} A closed-form
  noise-augmented Hessian, obtained as the expected GPTQ loss under an
  errors-in-variables model of upstream quantization noise.

  \item \textbf{Drop-in compatibility with GPTQ-family solvers.}
  CMPQ modifies only the Hessian matrix and leaves the column-wise
  solver and the output checkpoint format unchanged, so it inherits
  GPTQ's inference-time cost exactly while incurring only a moderate
  calibration-time overhead.

  \item \textbf{Empirical validation.} We evaluate CMPQ on three
  Qwen3-Base checkpoints (4B / 8B / 14B) at 3-bit and 2-bit weight-only
  quantization with six zero-shot tasks, and additionally on Qwen2.5-7B
  and Mistral-Nemo-12B. We also combine CMPQ with QuIP\#.
\end{itemize}

%% file: figures/gptq_blind.tex
%

\definecolor{gptqgray}{RGB}{75,75,75}
\definecolor{gptaqorange}{RGB}{200,80,40}
\definecolor{noisehaze}{RGB}{220,220,232}
\definecolor{soft}{RGB}{120,120,120}

\begin{tikzpicture}[
  refstar/.style={star, star points=5, star point ratio=2.3,
                  fill=green!45!black, draw=green!30!black,
                  inner sep=0pt, minimum size=8pt},
  obsdot/.style={circle, fill=gptaqorange, draw=gptaqorange!55!black,
                 line width=0.5pt, inner sep=0pt, minimum size=5.5pt}
]
  \fill[noisehaze, opacity=0.85] (3.0,2.2) ellipse [x radius=2.0, y radius=1.4, rotate=-15];
  \fill[noisehaze, opacity=0.6]  (3.0,2.2) ellipse [x radius=1.4, y radius=1.0, rotate=-15];

  \draw[->, soft, thick] (-0.1,-0.15) -- (6.0,-0.15)
    node[below left, black, xshift=2pt, font=\footnotesize] {$X^{(1)}$};
  \draw[->, soft, thick] (-0.1,-0.15) -- (-0.1,4.5)
    node[above, black, font=\footnotesize] {$X^{(2)}$};

  \node[refstar] (xref) at (3.0,2.2) {};
  \node[above=2pt, black, font=\small] at (xref) {$X^{\mathrm{ref}}$};

  \node[obsdot] (xobs) at (3.9,1.4) {};
  \node[below right=1pt, black, font=\small] at (xobs) {$X$};

  \draw[->, soft, dashed, line width=0.7pt] (xref) -- (xobs)
    node[midway, above right=-1pt, black, font=\footnotesize] {$\Delta X$};

  \draw[gptaqorange, line width=1pt]
    (xobs) ellipse [x radius=1.1, y radius=0.55, rotate=20];

  \node[anchor=south west, font=\footnotesize, gptaqorange] at (4.4,2)
    {$H = X X^{\!\top}$};

  \node[font=\footnotesize, soft] at (1.6,3.4) {noise neighborhood};
\end{tikzpicture}

%% file: figures/gptaq_drift-v2.tex
%

\definecolor{gptaqorange}{RGB}{200,80,40}
\definecolor{cmpqblue}{RGB}{40,90,170}
\definecolor{soft}{RGB}{120,120,120}

\begin{tikzpicture}[
  qmark/.style={circle, fill=gptaqorange, draw=gptaqorange!55!black,
                line width=0.5pt, inner sep=0pt, minimum size=5pt},
  tmark/.style={diamond, fill=cmpqblue!85, draw=cmpqblue!50!black,
                line width=0.5pt, inner sep=0pt, minimum size=5pt}
]
  \draw[->, soft, thick] (-0.1,-0.15) -- (6.0,-0.15)
    node[below left, black, xshift=2pt, font=\footnotesize] {column $q$};
  \draw[->, soft, thick] (-0.1,-0.15) -- (-0.1,4.5)
    node[above, black, font=\footnotesize] {residual};

  \fill[gptaqorange!12]
    (0.00,1.40) -- (0.92,1.42) -- (1.83,1.46) -- (2.75,1.50) --
    (3.67,1.54) -- (4.58,1.56) -- (5.50,1.55)            
    -- (5.50,3.50) -- (4.58,2.60) -- (3.67,1.95) --      
    (2.75,1.55) -- (1.83,1.32) -- (0.92,1.20) -- (0.00,1.40) -- cycle;

  \draw[cmpqblue, dashed, line width=1.1pt]
    (0.00,1.40) .. controls (1.83,1.45) and (3.67,1.55) .. (5.50,1.55);
  \foreach \x/\y in {0.00/1.40, 0.92/1.42, 1.83/1.46, 2.75/1.50,
                     3.67/1.54, 4.58/1.56, 5.50/1.55} {
    \node[tmark] at (\x,\y) {};
  }

  \draw[gptaqorange, line width=1.1pt]
    (0.00,1.40) -- (0.92,1.20) -- (1.83,1.32) -- (2.75,1.55) --
    (3.67,1.95) -- (4.58,2.60) -- (5.50,3.50);
  \foreach \x/\y in {0.00/1.40, 0.92/1.20, 1.83/1.32, 2.75/1.55,
                     3.67/1.95, 4.58/2.60, 5.50/3.50} {
    \node[qmark] at (\x,\y) {};
  }

  \draw[<->, soft, line width=0.7pt] (5.50,1.62) -- (5.50,3.43);

  \node[anchor=south east, gptaqorange, font=\footnotesize]
    at (4.80,2.65) {actual $W^{(q)}_{i,q:}\Delta X_{q:}$};

  \node[anchor=north east, cmpqblue, font=\footnotesize]
    at (5.30,1.35) {target $W^{(0)}_{i,q:}\Delta X_{q:}$};

  \node[anchor=east, soft, font=\footnotesize, align=right]
    at (6.40,2.55) {missed\\correction};
\end{tikzpicture}

%% file: sections/related-work.tex
\section{Related Work}
\label{sec:related}

\textbf{Post-training weight quantization.}
Post-training quantization (PTQ)~\cite{gholami2022survey} compresses
pre-trained model weights to low-bit representations using only a small
calibration set, without the gradient updates required by
quantization-aware training. For weight-only PTQ of large language models,
two algorithmic families have come to dominate practice.
\emph{Hessian-based solvers}~\cite{frantar2022optimal,frantar2022gptq,nagel2020up,li2021brecq,dettmers2023spqr,shao2023omniquant,chee2023quip,tseng2024quip}
model the per-layer reconstruction error as a quadratic in $\delta W$ and
minimize it column by column using second-order information from the
calibration input.
\emph{Activation-aware scalers}~\cite{lin2024awq,xiao2023smoothquant,dettmers2022gpt3}
apply per-channel scale factors that migrate quantization difficulty
between weights and activations, protecting channels with disproportionately
large activation magnitudes.
CMPQ targets the first family; the activation-aware line is parallel to
and composable with the Hessian augmentation we propose.

\textbf{Hessian-based PTQ solvers.}
The dominant approach to finetuning-free weight-only PTQ frames each
layer as a reconstruction problem: minimize $\|WX - \hat W X\|_F^2$ subject
to a quantization grid, guided by the Hessian $H = XX^\top$ of the
calibration input.
OBQ~\cite{frantar2022optimal} laid the theoretical foundation by solving
this problem one weight at a time; GPTQ~\cite{frantar2022gptq} scaled it
to LLMs via a column-wise lazy update scheme that propagates each column's
quantization error to the remaining unquantized columns.
Subsequent work extends the reconstruction unit from individual layers to
blocks (BRECQ~\cite{li2021brecq}, OmniQuant~\cite{shao2023omniquant}),
pre-processes the Hessian for incoherence before rounding
(QuIP~\cite{chee2023quip}, QuIP\#~\cite{tseng2024quip}), or uses Hessian
sensitivity scores to retain outlier weights at higher precision
(SpQR~\cite{dettmers2023spqr}).
All of these methods treat the observed calibration input $X$ as ground
truth, without accounting for the upstream quantization noise it carries.
Rotation- and incoherence-based methods such as QuIP and QuIP\# act on
the rounding side, changing the basis and the codebook, but still round
with an LDLQ/GPTQ-style solver whose Hessian is computed from the
quantized chain; CMPQ's augmented Hessian can be passed to that solver
unchanged (Section~\ref{sec:exp-quip}).

\textbf{Direct upstream-noise compensation.}
GPTAQ~\cite{li2025gptaq} explicitly addresses the discrepancy between the
quantized calibration input $X$ and its FP16 reference $X^{\text{ref}}$ by
changing the per-layer objective to
$\|W X^{\text{ref}} - \hat W X\|_F^2$. Its column-wise solver attaches a
first-order correction in $\Delta X = X - X^{\text{ref}}$ on top of the
standard GPTQ update. However, because this correction depends on the
partially-quantized weights $W^{(q)}$ at each column step, the injected
residual drifts as more columns are quantized; \cite{li2026rethinking}
identifies this drift as the \emph{compensation-aware error} and proposes
a column-level fix that re-anchors the correction to the pre-quantization
weights $W^{(0)}$.
CMPQ takes a structurally different angle: rather than augmenting the
column update with a first-order term, we encode $\Delta X$ as second-order
information ($V = \Delta X\,\Delta X^\top$) and inject it into the Hessian
as anisotropic damping. The augmented Hessian depends only on input
statistics and is fixed before the column loop, so CMPQ inherits no
analogue of the compensation-aware drift. Because the two corrections act
at different places, they could in principle be combined.

%% file: sections/preliminaries.tex
\section{Preliminaries}
\label{sec:prelim}

\paragraph{GPTQ.}
Consider a single linear layer $Y = WX$, where
$W \in \mathbb{R}^{d_{\mathrm{out}} \times d_{\mathrm{in}}}$ and
$X \in \mathbb{R}^{d_{\mathrm{in}} \times N}$ collects activations from $N$
calibration tokens.
GPTQ~\cite{frantar2022gptq} finds the quantized weight matrix $\hat{W}$ by minimizing the per-layer
reconstruction error
\begin{equation}
\label{eq:gptq-obj}
\min_{\hat{W}} \; \|\delta W\, X\|_F^2, \qquad \delta W \;=\; W - \hat{W},
\end{equation}
with Hessian $H = X X^\top$.
The solver proceeds column by column: after quantizing column $q$ to
$\hat{w}_q = \mathrm{quant}(w_q)$, the quantization error
$e_q = w_q - \hat{w}_q$ is propagated to all remaining columns $k > q$ via
\begin{equation}
\label{eq:gptq-update}
w_k \;\leftarrow\; w_k \;-\; \frac{e_q}{[H^{-1}]_{qq}}\,[H^{-1}]_{qk},
\end{equation}
where each output row is processed independently and $H^{-1}$ is obtained via
Cholesky decomposition.

\paragraph{GPTAQ.}
GPTAQ~\cite{li2025gptaq} makes the quant-reference discrepancy explicit by changing the
per-layer objective to
\begin{equation}
\label{eq:gptaq-obj}
\min_{\hat{W}} \; \|W X^{\text{ref}} - \hat{W} X\|_F^2,
\end{equation}
so that the target uses clean FP16 activations.
GPTAQ derives a column-wise solver for this objective.
Denote by $W^{(q)}$ the weight matrix after columns $0,\ldots,q-1$ have been
quantized and compensated, and let $e_{i,q} = w^{(q)}_{i,q} - \hat{w}_{i,q}$
be the quantization error at column $q$ for row $i$.
The GPTAQ update for each remaining column $k > q$ is
\begin{equation}
\label{eq:gptaq-update}
w_{i,k} \;\leftarrow\; w_{i,k}
  - \frac{e_{i,q}}{[H^{-1}]_{qq}}\,[H^{-1}]_{qk}
  + \bigl[\,r_i\,X^\top H^{-1}\,\bigr]_k,
\end{equation}
where the first term is the standard GPTQ compensation
(Eq.~\ref{eq:gptq-update}) and the second is GPTAQ's first-order correction,
with residual
\begin{equation}
\label{eq:gptaq-residual}
r_i \;=\; -\,w^{(q)}_{i,q:}\,\Delta X_{q:} \;\in\; \mathbb{R}^{1\times N},
\end{equation}
where $\Delta X = X - X^{\text{ref}}$ is the gap between the quantized-chain
and FP16-chain activations.

%% file: sections/methodology.tex
\section{Methodology}
\label{sec:method}


\paragraph{Issue of GPTQ: cumulative upstream noise.}
GPTQ optimizes the per-layer reconstruction objective
(Eq.~\ref{eq:gptq-obj}) against a single observed input $X$.
However, layers are quantized sequentially, so $X$ is not the clean
full-precision activation but the output of the cumulative quantized chain.
Denoting by $X^{\text{ref}}$ the activation produced by the FP16 chain, we
define the \textbf{upstream noise}
\begin{equation}
\label{eq:delta-x}
\Delta X \;\equiv\; X - X^{\text{ref}} \;\in\; \mathbb{R}^{d_{\mathrm{in}} \times N},
\end{equation}
which accumulates with depth and grows rapidly at low bit-widths.
By treating $X$ as if it were $X^{\text{ref}}$, GPTQ computes its Hessian
$H = XX^\top$ from a corrupted input, silently ignoring the upstream noise.
At inference time, the input arriving at this layer is a \emph{different}
realization of the same upstream-noise process.
What we need is a $\hat W$ whose reconstruction error stays small
\emph{across} this noise-shaped neighborhood, not just at the calibration
snapshot.

\paragraph{Issue of GPTAQ: compensation drift.}
GPTAQ attempts to account for the noise via a first-order correction with
residual
\begin{equation}
r_i \;=\; -\,W^{(q)}_{i,q:}\,\Delta X_{q:}
\end{equation}
(Eq.~\ref{eq:gptaq-residual}), but $r_i$ depends on the running weight
state $W^{(q)}$.
For columns $q \ge 1$, the correction targets $W^{(q)}_{i,q:}\Delta X_{q:}$
instead of $W^{(0)}_{i,q:}\Delta X_{q:}$, as shown in Figure~\ref{fig:gptaq-drift}.
Such compensation-aware error compounds with the number of quantized
columns~\cite{li2026rethinking}.

\paragraph{Our approach.}
Our proposed CMPQ addresses both problems by incorporating the upstream noise
structure into the Hessian as a static damping term.
The damping term captures where upstream noise lies in input space,
fixing GPTQ's blindness to it.
Since the term depends only on calibration statistics, it also
sidesteps GPTAQ's compensation drift.

\subsection{Errors-in-Variables Formulation}
We formalize this neighborhood with an errors-in-variables (EIV) model,
treating the calibration input as subject to a zero-mean Gaussian perturbation
\begin{equation}
\label{eq:eiv-perturb}
X' \;=\; X + \xi,
\qquad \xi \sim \mathcal{N}\!\left(0,\;\lambda\,V\right),
\qquad V \;=\; \Delta X\,\Delta X^\top,
\end{equation}
where $\lambda \ge 0$ is a scalar weight.
The covariance $V$ is the sample second moment of the observed
upstream noise across the $N$ calibration tokens, which is the standard plug-in
estimator for the noise distribution under the zero-mean assumption.
Expanding the squared Frobenius norm and taking expectation, the cross term
vanishes by $\mathbb{E}[\xi]=0$ and the quadratic term reduces to a trace
against $\mathbb{E}[\xi\xi^\top] = \lambda V$, giving
\begin{align}
\mathbb{E}_{\xi}\bigl[\|\delta W\,X'\|_F^2\bigr]
&\;=\; \mathbb{E}_{\xi}\bigl[\|\delta W\,(X+\xi)\|_F^2\bigr] \nonumber \\
&\;=\; \|\delta W\,X\|_F^2
   \;+\; 2\,\bigl\langle \delta W\,X,\;\delta W\,\mathbb{E}[\xi]\bigr\rangle_F
   \;+\; \mathbb{E}_{\xi}\bigl[\|\delta W\,\xi\|_F^2\bigr] \nonumber \\
\label{eq:eiv-prop}
&\;=\; \|\delta W\,X\|_F^2 \;+\; \lambda\,\|\delta W\,\Delta X\|_F^2.
\end{align}
The full derivation is in Appendix~\ref{app:derivation}.
In Eq.~(\ref{eq:eiv-prop}), the first term $\|\delta W\,X\|_F^2$ is
GPTQ's standard reconstruction objective, while the second term
$\lambda\,\|\delta W\,\Delta X\|_F^2$ penalizes $\delta W$ along the
upstream-noise directions.

\subsection{Noise-Augmented Hessian}
\label{sec:noise-aug-hessian}

\paragraph{CMPQ objective.}
We define the CMPQ per-layer objective as minimizing the expected
reconstruction loss under the EIV perturbation. By
Eq.~(\ref{eq:eiv-prop}),
\begin{equation}
\label{eq:cmpq-obj}
\hat W^{\star} \;=\; \arg\min_{\hat{W}} \;
\mathbb{E}_{\xi}\bigl[\|\delta W\,X'\|_F^2\bigr]
\;=\; \arg\min_{\hat{W}} \;
\|\delta W\,X\|_F^2 \;+\; \lambda\,\|\delta W\,\Delta X\|_F^2,
\qquad \lambda \ge 0.
\end{equation}
The optimization yields a $\hat W^{\star}$ that reconstructs the observed
input while remaining robust to upstream-noise perturbations, with
$\lambda$ controlling the trade-off; at $\lambda = 0$, the objective
recovers vanilla GPTQ.
Rewriting each squared Frobenius norm in Eq.~(\ref{eq:cmpq-obj}) as a
quadratic form in $\delta W$ (see Appendix~\ref{app:derivation}),
Eq.~(\ref{eq:cmpq-obj}) reduces to
$\mathrm{tr}(\delta W\,\hat{H}\,\delta W^\top)$ with the
\textbf{noise-augmented Hessian} formed by adding the noise covariance
$V = \Delta X\,\Delta X^\top$ to the GPTQ Hessian $H = XX^\top$:
\begin{equation}
\label{eq:aug-hessian}
\hat{H} \;=\; H \;+\; \lambda\,V
\;\in\; \mathbb{R}^{d_{\mathrm{in}} \times d_{\mathrm{in}}}.
\end{equation}
The GPTQ column-wise solver (Eq.~\ref{eq:gptq-update}) is then applied
verbatim with $\hat{H}$ in place of $H$, requiring no other change to the
solver.
$\hat{H}$ depends only on the calibration statistics $X$ and
$X^{\text{ref}}$ and is computed \emph{once} before the column loop.

\paragraph{Two equivalent views.}
$\hat H$ admits two equivalent reinterpretations: (i) as the Gram matrix
$\tilde X \tilde X^\top$ of a widened calibration matrix
$\tilde X = [\,X \mid \sqrt{\lambda}\,\Delta X\,]$, a finite-sample realization
of the EIV prior; and (ii) as generalized Tikhonov on $H$ with anisotropic
operator $\Delta X^\top$, the direction-aware analogue of GPTQ's ordinary
$\epsilon I$ damping. See Appendix~\ref{app:derivation} for details.

\paragraph{Trace-normalized form.}
\label{sec:trace-norm}
To give $\lambda$ a uniform interpretation across layers, we normalize $V_\ell$
at each layer $\ell$ by the trace ratio
$r_\ell = \mathrm{tr}(H_\ell)/(\mathrm{tr}(V_\ell)+\varepsilon)$, yielding the
\textbf{trace-normalized noise-augmented Hessian}:
\begin{equation}
\label{eq:trace-norm-hessian}
\hat{H}_\ell \;=\; H_\ell \;+\; \lambda\,r_\ell\,V_\ell.
\end{equation}
This rescaling enforces $\mathrm{tr}(r_\ell V_\ell) = \mathrm{tr}(H_\ell)$ at
every layer, so a single global $\lambda$ controls the overall damping
strength uniformly across the network. We adopt Eq.~(\ref{eq:trace-norm-hessian})
as the default CMPQ Hessian.

\subsection{Full Algorithm}
\label{sec:algo}

The trace-normalized noise-augmented Hessian
(Sec.~\ref{sec:noise-aug-hessian}) translates into a \emph{drop-in}
modification of GPTQ, shown in Algorithm~\ref{alg:cmpq}:
(i) forming $\Delta X_\ell$ from a paired FP16-chain forward pass,
(ii) accumulating $V_\ell = \Delta X_\ell\,\Delta X_\ell^\top$, and
(iii) building the trace-rescaled $\hat{H}_\ell$ via
Eq.~(\ref{eq:trace-norm-hessian}). The Cholesky factorization and
per-column greedy quantization with error compensation carry over
verbatim from GPTQ, leaving the column-wise solver untouched.

\begin{algorithm}[h]
\small
\caption{CMPQ quantization of one Transformer block.}
\label{alg:cmpq}
\begin{algorithmic}[1]
\Require FP16 weights $\{W_\ell\}$; calibration inputs $X^{\text{ref}}_0$, $X_0$; damping $\lambda$; numerical floor $\varepsilon$.
\State Forward both the FP16 chain ($X^{\text{ref}}_0$) and Q chain ($X_0$) through the block, capturing $X^{\text{ref}}_\ell, X_\ell$ at each linear input.
\For{each linear $\ell$ in the block}
    \State $\Delta X_\ell \gets X_\ell - X^{\text{ref}}_\ell$ \hfill \Comment{\textbf{CMPQ}}
    \State $H_\ell \gets X_\ell X_\ell^\top$;\quad $V_\ell \gets \Delta X_\ell\,\Delta X_\ell^\top$ \hfill \Comment{\textbf{CMPQ}}
    \State $\hat{H}_\ell \gets H_\ell + \lambda\,\bigl(\mathrm{tr}(H_\ell)/(\mathrm{tr}(V_\ell)+\varepsilon)\bigr)\,V_\ell$ \hfill \Comment{\textbf{CMPQ: Eq.~\ref{eq:trace-norm-hessian}}}
    \State $L_\ell \gets \operatorname{Cholesky}(\hat{H}_\ell)$
    \For{each column $q$ of $W_\ell$}
        \State $\hat{W}_{\ell,:,q} \gets \operatorname{quant}(W_{\ell,:,q})$
        \State Propagate error to remaining columns via Eq.~(\ref{eq:gptq-update}).
    \EndFor
\EndFor
\State Forward $X_0$ through the block with quantized weights $\{\hat{W}_\ell\}$ to update the Q chain for the next block.
\State \Return $\{\hat{W}_\ell\}$.
\end{algorithmic}
\end{algorithm}

\paragraph{Overhead.}
CMPQ adds one FP16-chain forward pass per block and an extra
$\mathcal{O}(d_{\mathrm{in}}^2 N)$ accumulation for $V_\ell$ on top of GPTQ;
the dominant $\mathcal{O}(d_{\mathrm{in}}^3)$ Cholesky cost is unchanged
(Section~\ref{sec:exp-complexity}).

%% file: sections/experiments.tex
\section{Experiments}
\label{sec:experiments}

We evaluate CMPQ on three pretrained Qwen3-Base checkpoints across
three size points (4B, 8B, 14B) under 2-bit and 3-bit weight-only
quantization settings. Section~\ref{sec:exp-setup} describes the
setup and Section~\ref{sec:exp-main} the main results against three
baselines. Section~\ref{sec:exp-quip} combines CMPQ with QuIP\#,
Section~\ref{sec:exp-ablation} ablates its components and damping
structure, Section~\ref{sec:exp-lambda} examines sensitivity to
$\lambda$, and Section~\ref{sec:exp-complexity} reports the
calibration cost.

\begin{table*}[t]
  \centering
  \footnotesize
  \begin{tabular}{ll cc cccccc c}
  \toprule
  & & \multicolumn{2}{c}{PPL $\downarrow$}
    & \multicolumn{6}{c}{Zero-shot Acc.\ (\%) $\uparrow$}
    & \\
  \cmidrule(lr){3-4} \cmidrule(lr){5-10}
  Model & Method & WT2 & C4 & ARC-c & ARC-e & BoolQ & HS & PIQA & WG & Avg$\uparrow$ \\
  \midrule
  \multirow{5}{*}{Qwen3-4B}
    & BF16                          & 7.90               & 11.52              & 51.45              & 76.01              & 82.72              & 73.70              & 77.80              & 70.96              & 72.11              \\
    \cmidrule(lr){2-11}
    & GPTQ                          & 328.78             & 199.43             & 23.89              & 25.46              & 44.95              & 27.29              & 51.47              & 52.88              & 37.66              \\
    & AWQ                           & 7.4e4              & 4.9e4              & 25.85              & 26.01              & 44.98              & 26.37              & 50.16              & 51.30              & 37.45              \\
    & GPTAQ                         & \underline{36.91}  & \underline{33.60}  & \underline{26.37}  & \underline{44.11}  & \underline{62.32}  & \underline{37.86}  & \underline{63.22}  & \underline{54.06}  & \underline{47.99}  \\
    & CMPQ                          & \textbf{26.74}     & \textbf{26.31}     & \textbf{31.83}     & \textbf{55.81}     & \textbf{64.98}     & \textbf{49.50}     & \textbf{67.36}     & \textbf{58.96}     & \textbf{54.74}     \\
  \midrule
  \addlinespace[2pt]
  \multirow{5}{*}{Qwen3-8B}
    & BF16                          & 7.00               & 10.37              & 56.57              & 80.22              & 83.12              & 78.68              & 79.43              & 71.59              & 74.93              \\
    \cmidrule(lr){2-11}
    & GPTQ                          & 114.0              & 66.14              & 25.00              & 27.40              & 47.52              & 31.92              & 54.19              & 49.80              & 39.30              \\
    & AWQ                           & 1.2e4              & 5.2e3              & 24.32              & 28.03              & 42.51              & 27.88              & 52.45              & 49.57              & 37.46              \\
    & GPTAQ                         & \underline{27.24}  & \underline{25.21}  & \underline{29.52}  & \underline{54.46}  & \underline{62.20}  & \underline{43.64}  & \underline{66.10}  & \underline{56.35}  & \underline{52.05}  \\
    & CMPQ                          & \textbf{23.01}     & \textbf{22.37}     & \textbf{36.60}     & \textbf{63.09}     & \textbf{65.69}     & \textbf{53.50}     & \textbf{69.86}     & \textbf{59.75}     & \textbf{58.08}     \\
  \midrule
  \addlinespace[2pt]
  \multirow{5}{*}{Qwen3-14B}
    & BF16                          & 6.38               & 9.63               & 59.04              & 82.15              & 86.64              & 81.36              & 82.05              & 74.59              & 77.64              \\
    \cmidrule(lr){2-11}
    & GPTQ                          & 46.22              & 34.14              & 25.26              & 32.37              & 54.04              & 38.94              & 55.66              & 51.54              & 42.97              \\
    & AWQ                           & 858.83             & 339.54             & 25.43              & 28.45              & 44.43              & 30.10              & 51.85              & 48.86              & 38.19              \\
    & GPTAQ                         & \textbf{14.44}     & \textbf{16.34}     & \underline{40.70}  & \underline{65.53}  & \underline{67.43}  & \underline{60.75}  & \underline{72.85}  & \underline{67.40}  & \underline{62.45}  \\
    & CMPQ                          & \underline{14.90}  & \underline{16.92}  & \textbf{43.43}     & \textbf{70.75}     & \textbf{72.69}     & \textbf{63.09}     & \textbf{73.83}     & \textbf{70.32}     & \textbf{65.68}     \\
  \bottomrule
  \end{tabular}
  \caption{Main results at \textbf{2-bit} weight-only quantization.
  Within each model group, best per column is in \textbf{bold} and
  second-best is \underline{underlined}; the unquantized checkpoint
  is shown as a reference and
  excluded from the ranking. $\lambda$ selection and metrics follow
  Section~\ref{sec:exp-setup}; the full $\lambda$ sweep is in
  Figure~\ref{fig:lambda-sweep}.}
  \label{tab:main-w2}
  \end{table*}

  \begin{table*}[t]
  \centering
  \footnotesize
  \begin{tabular}{ll cc cccccc c}
  \toprule
  & & \multicolumn{2}{c}{PPL $\downarrow$}
    & \multicolumn{6}{c}{Zero-shot Acc.\ (\%) $\uparrow$}
    & \\
  \cmidrule(lr){3-4} \cmidrule(lr){5-10}
  Model & Method & WT2 & C4 & ARC-c & ARC-e & BoolQ & HS & PIQA & WG & Avg$\uparrow$ \\
  \midrule
  \multirow{5}{*}{Qwen3-4B}
    & BF16                          & 7.90               & 11.52              & 51.45              & 76.01              & 82.72              & 73.70              & 77.80              & 70.96              & 72.11              \\
    \cmidrule(lr){2-11}
    & GPTQ                          & 10.12              & 13.60              & 45.39              & 66.54              & 60.52              & \underline{67.70}  & 74.97              & 65.11              & 63.37              \\
    & AWQ                           & 10.86              & 14.67              & 40.53              & 59.89              & \underline{72.39}  & 67.58              & 73.72              & 65.35              & 63.24              \\
    & GPTAQ                         & \textbf{9.54}      & \textbf{13.23}     & \textbf{48.29}     & \textbf{76.60}     & \textbf{82.26}     & 66.78              & \textbf{75.79}     & \underline{67.32}  & \textbf{69.51}     \\
    & CMPQ                          & \underline{9.58}   & \underline{13.30}  & \underline{47.95}  & \underline{71.93}  & 70.67              & \textbf{68.10}     & \underline{75.24}  & \textbf{68.90}     & \underline{67.13}  \\
  \midrule
  \addlinespace[2pt]
  \multirow{5}{*}{Qwen3-8B}
    & BF16                          & 7.00               & 10.37              & 56.57              & 80.22              & 83.12              & 78.68              & 79.43              & 71.59              & 74.93              \\
    \cmidrule(lr){2-11}
    & GPTQ                          & 8.63               & 11.83              & 49.57              & 72.98              & 80.18              & 74.54              & 77.37              & 66.54              & 70.20              \\
    & AWQ                           & 8.93               & 12.12              & 53.50              & 76.94              & 83.15              & \textbf{74.84}     & 77.48              & 67.48              & 72.23              \\
    & GPTAQ                         & \textbf{8.25}      & \textbf{11.62}     & \textbf{55.89}     & \textbf{80.72}     & \textbf{85.11}     & 73.16              & \underline{78.94}  & \textbf{71.74}     & \textbf{74.26}     \\
    & CMPQ                          & \underline{8.38}   & \underline{11.69}  & \underline{55.38}  & \underline{79.46}  & \underline{84.46}  & \underline{74.67}  & \textbf{79.05}     & \underline{70.88}  & \underline{73.98}  \\
  \midrule
  \addlinespace[2pt]
  \multirow{5}{*}{Qwen3-14B}
    & BF16                          & 6.38               & 9.63               & 59.04              & 82.15              & 86.64              & 81.36              & 82.05              & 74.59              & 77.64              \\
    \cmidrule(lr){2-11}
    & GPTQ                          & 7.76               & 10.82              & 56.57              & 80.22              & 81.65              & \underline{77.89}  & 79.27              & 69.53              & 74.19              \\
    & AWQ                           & 7.61               & 10.84              & 54.44              & 75.72              & 79.82              & \textbf{78.45}     & 78.67              & 70.32              & 72.90              \\
    & GPTAQ                         & \textbf{7.19}      & \textbf{10.51}     & \underline{57.59}  & \underline{83.08}  & \textbf{86.88}     & 77.19              & \underline{80.09}  & \textbf{76.16}     & \underline{76.83}  \\
    & CMPQ                          & \underline{7.24}   & \underline{10.56}  & \textbf{60.41}     & \textbf{83.54}     & \underline{85.14}  & 77.83              & \textbf{80.52}     & \underline{74.66}  & \textbf{77.02}     \\
  \bottomrule
  \end{tabular}
  \caption{Main results at \textbf{3-bit} weight-only quantization.
  Columns and conventions match Table~\ref{tab:main-w2}.}
  \label{tab:main-w3}
  \end{table*}

\subsection{Experimental Setup}
\label{sec:exp-setup}

\textbf{Models.}
We use three pretrained Qwen3-Base
checkpoints~\cite{yang2025qwen3} across three size points: Qwen3-4B,
Qwen3-8B, and Qwen3-14B. All models are evaluated at their released
BF16 precision before quantization; instruction-tuned and chat
variants are excluded. Appendix~\ref{app:families} additionally
reports Qwen2.5-7B and Mistral-Nemo-12B.

\textbf{Baselines.}
We compare against three baselines (see Section~\ref{sec:prelim}):
(i) \textbf{GPTQ}~\cite{frantar2022gptq}, the vanilla column-wise
Hessian-based solver;
(ii) \textbf{AWQ}~\cite{lin2024awq}, activation-aware per-channel
scaling;
(iii) \textbf{GPTAQ}~\cite{li2025gptaq}, which augments GPTQ with a
first-order correction that compensates for upstream activation
quan\-ti\-za\-tion noise. For all baselines, we follow the default
settings from their original papers, except for AWQ: we use the
\texttt{llm\_compressor} package because its original implementation
is numerically unstable on Qwen3 models.

\textbf{CMPQ configuration.}
All CMPQ experiments use the trace-normalized Hessian
$\hat{H}_\ell = H_\ell + \lambda\,r_\ell\,V_\ell$ with
$r_\ell = \mathrm{tr}(H_\ell)/(\mathrm{tr}(V_\ell)+\varepsilon)$
(Eq.~\ref{eq:trace-norm-hessian}), and the same per-group MSE-optimal
clip search as in GPTAQ. We experiment with four damping
strengths $\lambda \in \{0.01, 0.1, 0.5, 1\}$ for every model/bit-width
combination and select, per model and bit-width, the $\lambda$ with the
lowest WikiText-2 perplexity; the zero-shot tasks are used only for
evaluation and never for selection.

\textbf{Implementation details.}
We implement CMPQ on top of Hugging
Face Transformers~\cite{wolf2019huggingface} and
PyTorch~\cite{paszke2019pytorch}. Quantization is symmetric and weight-only, with group size
$128$, applied at $3$- and $2$-bit precision; activations are kept at
the released BF16 precision. For calibration, we use $128$ sequences of length $2048$
drawn from the C4~\cite{raffel2020exploring} train split with a
fixed seed ($42$; Appendix~\ref{app:seeds} varies it), with act-order column permutation throughout; the
full-precision and quantized forward chains required by CMPQ
(Section~\ref{sec:noise-aug-hessian}) share this calibration set.

\textbf{Evaluation.}
Perplexity is reported on WikiText-2~\cite{merity2016pointer}
(concatenated end-to-end) and on $256$ randomly-sampled $2048$-token
sequences from C4 validation. We report zero-shot accuracy with
\texttt{lm-evaluation-harness}~\cite{gao2021framework} on six
tasks (ARC-Easy, ARC-Challenge, HellaSwag, PIQA, WinoGrande, BoolQ),
using \texttt{acc\_norm} for ARC/HellaSwag/PIQA and \texttt{acc} for
BoolQ/WinoGrande, plus the six-task average (Avg).

\subsection{Main Results}
\label{sec:exp-main}

Across both bit-widths and all three models, CMPQ improves over
GPTQ on both perplexities and on the zero-shot average, with the
largest gains at 2-bit.

\paragraph{2-bit.}
At 2-bit (Table~\ref{tab:main-w2}), the most aggressive regime, both
GPTQ and AWQ collapse to near-random ZS averages (around $37$--$43$
on all three models), indicating that neither column-wise
compensation nor activation-aware scaling alone absorbs the
upstream noise at this bit-width. CMPQ holds $54.74$ / $58.08$ /
$65.68$ on Qwen3-4B / 8B / 14B, gains of $+17.08$ / $+18.78$ /
$+22.71$ over GPTQ. CMPQ also pulls ahead of GPTAQ on every model
($+6.75$ / $+6.03$ / $+3.23$), winning every zero-shot column on
all three. Translated into recovery relative to the BF16 reference,
CMPQ retains roughly $76\%$ / $78\%$ / $85\%$ of the BF16 zero-shot
average on the three models, versus $52\%$--$55\%$ for GPTQ and
$67\%$--$80\%$ for GPTAQ, leaving CMPQ within $12$--$17$
zero-shot points of BF16. GPTAQ also recovers much of the GPTQ
collapse, but CMPQ recovers more.

\paragraph{3-bit.}
At 3-bit (Table~\ref{tab:main-w3}) CMPQ delivers clear
improvements over GPTQ on every model: $+3.76$ on Qwen3-4B, $+3.78$
on Qwen3-8B, and $+2.83$ on Qwen3-14B, together with lower
perplexity on all three. Against GPTAQ, results are mixed: CMPQ leads
by $0.19$ on Qwen3-14B and trails by $0.28$ on Qwen3-8B and $2.38$ on
Qwen3-4B, while GPTAQ's WikiText-2 perplexity is lower by at most
$0.13$.

\paragraph{Calibration-seed variability.}
All main-table numbers come from a single calibration seed. Repeating
the 2-bit comparison with GPTAQ over four seeds
(Appendix~\ref{app:seeds}, Table~\ref{tab:seeds}), CMPQ keeps the
higher mean zero-shot accuracy on both Qwen3-8B and Qwen3-14B.

\paragraph{Takeaways.}
CMPQ improves over GPTQ at both bit-widths; its advantage over GPTAQ
is clear at 2-bit and mixed at 3-bit.

\subsection{Composing with Rounding-Side Methods}
\label{sec:exp-quip}

CMPQ changes only the Hessian handed to the solver, so it applies to
any method whose rounding step is driven by a calibration Hessian.
QuIP\#~\cite{tseng2024quip} is such a method: it applies randomized
Hadamard incoherence processing and rounds onto E8P lattice codebooks
with BlockLDLQ, an LDL-based generalization of the GPTQ update whose
Hessian is computed on the quantized chain. We reimplement QuIP\#
without its fine-tuning stage in our pipeline and, for QuIP\# + CMPQ,
pass $\hat H = H + \lambda r V$ to the same BlockLDLQ step
(Table~\ref{tab:quip-cmpq}).
\begin{table*}[t]
\centering
\footnotesize
\begin{tabular}{ll cc cccccc c}
\toprule
& & \multicolumn{2}{c}{PPL $\downarrow$} & \multicolumn{6}{c}{Zero-shot Acc.\ (\%) $\uparrow$} & \\
\cmidrule(lr){3-4} \cmidrule(lr){5-10}
Setting & Method & WT2 & C4 & ARC-c & ARC-e & BoolQ & HS & PIQA & WG & Avg$\uparrow$ \\
\midrule
\multirow{3}{*}{\shortstack[l]{Qwen3-8B\\2-bit}}
  & CMPQ               & 23.01 & 22.37 & 36.60 & 63.09 & 65.69 & 53.50 & 69.86 & 59.75 & 58.08 \\
  & QuIP\# (no FT)     & 9.33 & \textbf{12.71} & 46.93 & 73.44 & 79.76 & \textbf{69.86} & \textbf{77.48} & 68.27 & 69.29 \\
  & QuIP\# + CMPQ      & \textbf{9.18} & 12.72 & \textbf{51.45} & \textbf{79.04} & \textbf{82.51} & 68.64 & 76.77 & \textbf{69.22} & \textbf{71.27} \\
\midrule
\multirow{3}{*}{\shortstack[l]{Qwen3-8B\\3-bit}}
  & CMPQ               & 8.38 & 11.69 & 55.38 & 79.46 & \textbf{84.46} & 74.67 & \textbf{79.05} & 70.88 & 73.98 \\
  & QuIP\# (no FT)     & 7.48 & \textbf{10.96} & \textbf{58.45} & \textbf{82.49} & 82.78 & \textbf{75.93} & 78.89 & \textbf{72.45} & \textbf{75.17} \\
  & QuIP\# + CMPQ      & \textbf{7.46} & 10.97 & 56.48 & 82.03 & 83.61 & 75.52 & 78.89 & 71.90 & 74.74 \\
\midrule
\multirow{3}{*}{\shortstack[l]{Qwen3-14B\\2-bit}}
  & CMPQ               & 14.90 & 16.92 & 43.43 & 70.75 & 72.69 & 63.09 & 73.83 & 70.32 & 65.68 \\
  & QuIP\# (no FT)     & 8.25 & 11.59 & 52.56 & \textbf{79.63} & \textbf{85.05} & 73.77 & 77.91 & 69.85 & 73.13 \\
  & QuIP\# + CMPQ      & \textbf{8.14} & \textbf{11.42} & \textbf{53.33} & 79.04 & 83.70 & \textbf{74.49} & \textbf{78.62} & \textbf{70.80} & \textbf{73.33} \\
\midrule
\multirow{3}{*}{\shortstack[l]{Qwen3-14B\\3-bit}}
  & CMPQ               & 7.24 & 10.56 & \textbf{60.41} & \textbf{83.54} & 85.14 & 77.83 & \textbf{80.52} & 74.66 & \textbf{77.02} \\
  & QuIP\# (no FT)     & 6.81 & 10.10 & 56.91 & 80.98 & 85.84 & \textbf{79.35} & 80.20 & 73.48 & 76.13 \\
  & QuIP\# + CMPQ      & \textbf{6.75} & \textbf{10.06} & 57.34 & 82.03 & \textbf{86.73} & 79.19 & 80.47 & \textbf{75.30} & 76.84 \\
\bottomrule
\end{tabular}
\caption{CMPQ combined with QuIP\# (no fine-tuning). Best per column
within each setting is in \textbf{bold}.}
\label{tab:quip-cmpq}
\end{table*}
Adding CMPQ lowers WikiText-2 perplexity in all four settings and
improves the zero-shot average in three of them, most on Qwen3-8B at
2-bit.

\subsection{Ablation Study}
\label{sec:exp-ablation}

Table~\ref{tab:ablation} decomposes CMPQ's contribution on Qwen3-14B
by adding one component at a time on top of GPTQ.
\begin{table*}[t]
\centering
\footnotesize
\begin{tabular}{l cc cccccc c}
\toprule
& \multicolumn{2}{c}{PPL $\downarrow$}
  & \multicolumn{6}{c}{Zero-shot Acc.\ (\%) $\uparrow$}
  & \\
\cmidrule(lr){2-3} \cmidrule(lr){4-9}
Recipe & WT2 & C4 & ARC-c & ARC-e & BoolQ & HS & PIQA & WG & Avg$\uparrow$ \\
\midrule
\multicolumn{10}{l}{\emph{3-bit}} \\
GPTQ                              & 7.76              & 10.82             & 56.57             & 80.22             & \underline{81.65} & \textbf{77.89}    & 79.27             & 69.53             & 74.19             \\
\;\;$+$ $V$ (no trace)            & \underline{7.25}  & \underline{10.57} & \underline{59.73} & \textbf{83.67}    & 81.56             & 77.38             & \textbf{80.85}    & \underline{74.27} & \underline{76.24} \\
\;\;$+$ $V$ $+$ trace (CMPQ)      & \textbf{7.24}     & \textbf{10.56}    & \textbf{60.41}    & \underline{83.54} & \textbf{85.14}    & \underline{77.83} & \underline{80.52} & \textbf{74.66}    & \textbf{77.02}    \\
\midrule
\multicolumn{10}{l}{\emph{2-bit}} \\
GPTQ                              & 46.22             & 34.14             & 25.26             & 32.37             & 54.04             & 38.94             & 55.66             & 51.54             & 42.97             \\
\;\;$+$ $V$ (no trace)            & \textbf{14.62}    & \underline{17.01} & \underline{43.09} & \underline{67.72} & \underline{68.90} & \textbf{63.33}    & \underline{73.07} & \underline{69.38} & \underline{64.25} \\
\;\;$+$ $V$ $+$ trace (CMPQ)      & \underline{14.90} & \textbf{16.92}    & \textbf{43.43}    & \textbf{70.75}    & \textbf{72.69}    & \underline{63.09} & \textbf{73.83}    & \textbf{70.32}    & \textbf{65.68}    \\
\bottomrule
\end{tabular}
\caption{Component ablation on Qwen3-14B at 3- and 2-bit.
Best is in \textbf{bold} and
second-best is \underline{underlined}.}
\label{tab:ablation}
\end{table*}
On Qwen3-14B, adding the $V$-augmented Hessian on top of GPTQ lifts
the zero-shot average by $+2.05$ at 3-bit ($74.19 \to 76.24$) and
substantially recovers the 2-bit collapse ($42.97 \to 64.25$,
$+21.28$). Adding trace normalization on top further improves the
zero-shot average at both bit-widths ($+0.78$ at 3-bit, $+1.43$ at
2-bit) and changes perplexity by at most $0.3$ WT2 in either
direction. The $V$ term carries most of the gain; trace normalization
adds a smaller zero-shot improvement at both bit-widths.

\paragraph{Damping structure.}
To isolate the role of the covariance structure, we replace $\lambda r V$
by isotropic ($\lambda\,\mathrm{tr}(H)/d \cdot I$) or diagonal
($\lambda r\,\mathrm{diag}(V)$) damping of the same trace on Qwen3-8B at
2-bit (Appendix~\ref{app:damping}, Table~\ref{tab:damping}). Full $V$
gives the lowest WikiText-2 and C4 perplexity at every
$\lambda \in \{0.01, 0.1, 1\}$ and is the only variant whose perplexity
and zero-shot accuracy stay stable as $\lambda$ grows; isotropic and
diagonal damping degrade sharply at larger $\lambda$.

\subsection{$\lambda$ Sensitivity}
\label{sec:exp-lambda}

\begin{figure}[t]
  \centering
  \input{figures/lambda_sensitivity}
  \caption{Zero-shot accuracy vs.\ damping strength $\lambda$ for CMPQ
  across the three Qwen3 models, at 3-bit (left) and 2-bit (right).}
  \label{fig:lambda-sweep}
  \end{figure}
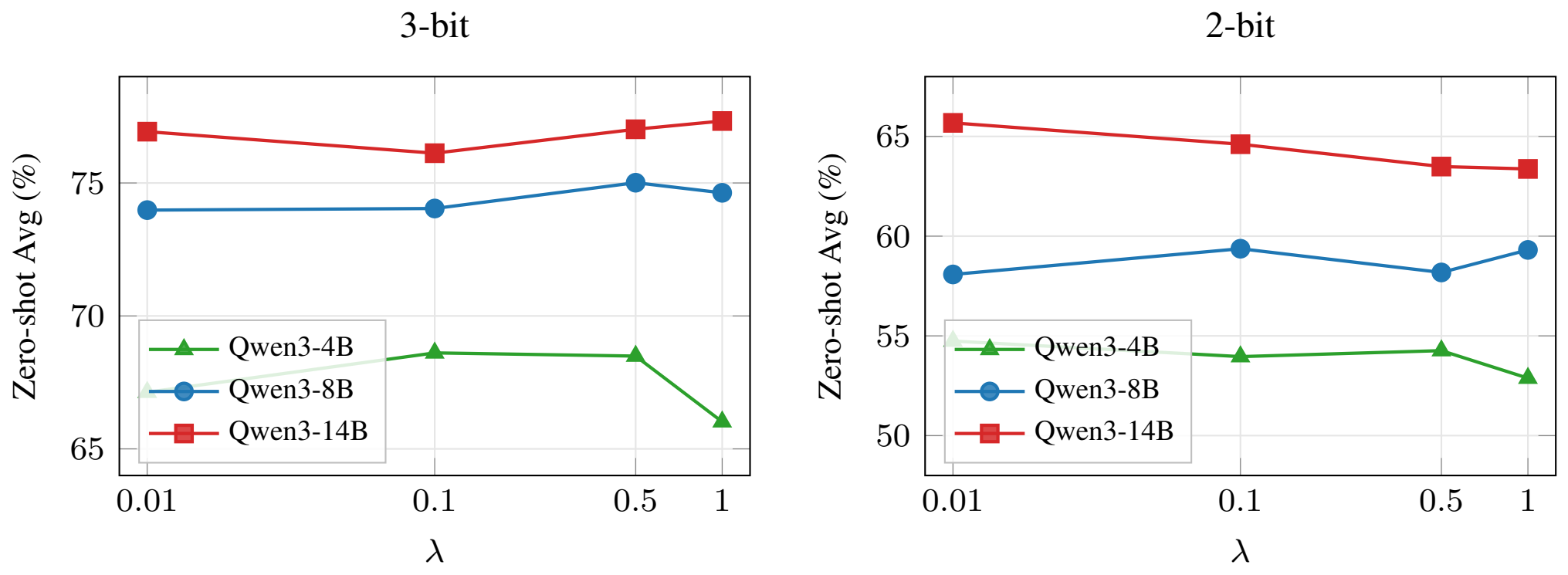

Figure~\ref{fig:lambda-sweep} reports the full $\lambda$ sweep for
CMPQ across our three Qwen3 models at both bit-widths.
Across $\lambda \in [0.01, 1]$, the zero-shot average of each model
varies by $1.0$--$2.6$ points at 3-bit and $1.3$--$2.3$ points at
2-bit. Selecting
$\lambda$ by WikiText-2 perplexity picks $\lambda{=}0.01$ in five of
the six model/bit-width cells (the exception is Qwen3-14B at 3-bit,
$\lambda{=}0.5$), and the selected value is within $1.5$ zero-shot
points of the best value in hindsight in every cell.

\subsection{Computational and Memory Overhead}
\label{sec:exp-complexity}

CMPQ adds work only at calibration time. Its output has the same
per-group quantization parameters as GPTQ at the same bit-width, with
no auxiliary state, so any kernel that loads a GPTQ checkpoint loads a
CMPQ checkpoint without modification, and inference-time cost is
identical.

The calibration overhead is moderate. Forming
$V_\ell = \Delta X_\ell\,\Delta X_\ell^\top$ costs
$\mathcal{O}(d_{\mathrm{in}}^2 N)$, the same order as
$H_\ell = X_\ell X_\ell^\top$; the addition
$H_\ell + \lambda\,r_\ell V_\ell$ is $\mathcal{O}(d_{\mathrm{in}}^2)$
and the trace ratio $\mathcal{O}(d_{\mathrm{in}})$, leaving the
dominant $\mathcal{O}(d_{\mathrm{in}}^3)$ Cholesky factorization
unchanged. Memory adds one $d_{\mathrm{in}} \times N$ activation
buffer for $\Delta X_\ell$ (or equivalently $X^{\text{ref}}_\ell$);
alternatively, the widened-calibration form $\tilde{X}_\ell =
[X_\ell \mid \sqrt{\lambda}\,\Delta X_\ell]$ can be streamed in
place of $X_\ell$ to avoid the explicit buffer. Capturing
$X^{\text{ref}}_\ell$ requires one extra forward pass per block
through the full-precision chain, shared across all linear
projections in the block. Table~\ref{tab:wallclock} (Appendix~\ref{app:efficiency}) reports
measurements at 2-bit on a single A100. CMPQ needs modestly more peak
memory than GPTQ and GPTAQ, and its quantization wall-clock time is
within the run-to-run variation of the three methods.

\section{Conclusion}
\label{sec:conclusion}

We presented CMPQ, a weight-only post-training quantization method
that augments the GPTQ Hessian with a static covariance term
capturing the cumulative upstream quantization noise. Computing this
term once from calibration data avoids the column-wise drift inherent
in first-order corrections like GPTAQ, while preserving the GPTQ
checkpoint format and inference path. Empirically, CMPQ improves
over GPTQ on every model and bit-width tested, and over GPTAQ in
zero-shot accuracy on every model at 2-bit, with the largest gains at
2-bit.

\section{Limitations}
\label{sec:limitations}

We note two limitations.
(i) \emph{Zero-mean assumption.} The covariance estimator $V$ is
faithful only when the upstream noise has zero mean. On the Qwen3
models, the share of $\mathrm{tr}(V)$ explained by a nonzero noise mean
has a per-layer median of $0.06$--$0.09$ and a maximum of
$0.29$--$0.37$ at 2-bit.
(ii) \emph{Scope of comparison.} Our combination with
QuIP\#~\cite{tseng2024quip} omits its fine-tuning stage and covers only two
model sizes.

%% file: figures/lambda_sensitivity.tex
%

\definecolor{model8b}{RGB}{31,119,180}
\definecolor{model14b}{RGB}{214,39,40}
\definecolor{modelllama}{RGB}{44,160,44}

\begin{tikzpicture}
\begin{groupplot}[%
  group style={group size=2 by 1, horizontal sep=1.5cm},
  width=0.5\linewidth, height=5cm,
  xmode=log, log basis x=10,
  xmin=0.008, xmax=1.25,
  xlabel={$\lambda$},
  ylabel={Zero-shot Avg (\%)},
  xtick={0.01, 0.1, 0.5, 1},
  xticklabels={$0.01$, $0.1$, $0.5$, $1$},
  grid=both, grid style={gray!20},
  minor grid style={gray!10},
  tick label style={font=\footnotesize},
  label style={font=\footnotesize},
  title style={font=\small},
  legend cell align=left,
  legend style={font=\scriptsize, draw=gray!50,
                fill=white, fill opacity=0.85, text opacity=1,
                inner sep=2pt},
]

\nextgroupplot[title={3-bit}, ymin=64, ymax=79,
               legend pos=south west]
\addplot[color=modelllama, mark=triangle*,thick] coordinates {
  (0.01, 67.13) (0.1, 68.61) (0.5, 68.49) (1, 66.02)};
\addlegendentry{Qwen3-4B}
\addplot[color=model8b,    mark=*,        thick] coordinates {
  (0.01, 73.98) (0.1, 74.04) (0.5, 75.01) (1, 74.63)};
\addlegendentry{Qwen3-8B}
\addplot[color=model14b,   mark=square*,  thick] coordinates {
  (0.01, 76.93) (0.1, 76.12) (0.5, 77.02) (1, 77.33)};
\addlegendentry{Qwen3-14B}

\nextgroupplot[title={2-bit}, ymin=48, ymax=68,
               legend pos=south west]
\addplot[color=modelllama, mark=triangle*,thick] coordinates {
  (0.01, 54.74) (0.1, 53.96) (0.5, 54.26) (1, 52.88)};
\addlegendentry{Qwen3-4B}
\addplot[color=model8b,    mark=*,        thick] coordinates {
  (0.01, 58.08) (0.1, 59.37) (0.5, 58.18) (1, 59.31)};
\addlegendentry{Qwen3-8B}
\addplot[color=model14b,   mark=square*,  thick] coordinates {
  (0.01, 65.68) (0.1, 64.61) (0.5, 63.49) (1, 63.37)};
\addlegendentry{Qwen3-14B}

\end{groupplot}
\end{tikzpicture}

%% file: sections/appendix-derivation.tex
\section{Derivation of the Noise-Augmented Hessian}
\label{app:derivation}

This appendix collects the technical material behind CMPQ. We (i) derive the EIV expansion in Eq.~(\ref{eq:eiv-prop}) and the resulting noise-augmented Hessian $\hat H = H + \lambda V$ (Eq.~\ref{eq:aug-hessian}); (ii) show that $\hat H$ is equivalently obtained by running vanilla GPTQ on a widened calibration matrix, so CMPQ requires \emph{no} modification to the solver; and (iii) place $\hat H$ in the framework of generalized Tikhonov regularization, recovering GPTQ's $\epsilon$-floor as the isotropic special case.

\subsection{Detailed expansion of the expected loss}

Starting from the perturbation model in Eq.~(\ref{eq:eiv-perturb}),
$X' = X + \xi$ with $\xi \sim \mathcal{N}(0,\,\lambda V)$ and
$V = \Delta X\,\Delta X^\top$, expand the squared Frobenius norm:
\begin{align}
\|\delta W\,X'\|_F^2
&\;=\; \|\delta W\,X + \delta W\,\xi\|_F^2 \nonumber \\
&\;=\; \|\delta W\,X\|_F^2
   \;+\; 2\,\bigl\langle \delta W\,X,\;\delta W\,\xi\bigr\rangle_F
   \;+\; \|\delta W\,\xi\|_F^2.
\end{align}
Take expectation over $\xi$. The cross term vanishes by linearity of the
inner product and $\mathbb{E}[\xi]=0$:
\begin{align}
\mathbb{E}_{\xi}\bigl[2\,\bigl\langle \delta W\,X,\;\delta W\,\xi\bigr\rangle_F\bigr]
\;=\; 2\,\bigl\langle \delta W\,X,\;\delta W\,\mathbb{E}[\xi]\bigr\rangle_F
\;=\; 0.
\end{align}
The remaining quadratic term simplifies via the cyclic trace identity
$\|A\|_F^2 = \mathrm{tr}(A^\top A) = \mathrm{tr}(A A^\top)$:
\begin{align}
\mathbb{E}_{\xi}\bigl[\|\delta W\,\xi\|_F^2\bigr]
&\;=\; \mathbb{E}_{\xi}\bigl[\mathrm{tr}\bigl(\delta W\,\xi\,\xi^\top\,\delta W^\top\bigr)\bigr] \nonumber \\
&\;=\; \mathrm{tr}\bigl(\delta W\;\mathbb{E}[\xi\,\xi^\top]\;\delta W^\top\bigr) \nonumber \\
&\;=\; \mathrm{tr}\bigl(\delta W \cdot \lambda V \cdot \delta W^\top\bigr) \nonumber \\
&\;=\; \lambda\,\mathrm{tr}\bigl(\delta W\,\Delta X\,\Delta X^\top\,\delta W^\top\bigr) \nonumber \\
&\;=\; \lambda\,\|\delta W\,\Delta X\|_F^2,
\end{align}
where the third line uses $\mathbb{E}[\xi\xi^\top] = \lambda V$ from
Eq.~(\ref{eq:eiv-perturb}) and the fourth substitutes $V = \Delta X\Delta X^\top$.
Combining:
\begin{equation}
\mathbb{E}_{\xi}\bigl[\|\delta W\,X'\|_F^2\bigr]
\;=\; \|\delta W\,X\|_F^2 \;+\; \lambda\,\|\delta W\,\Delta X\|_F^2,
\end{equation}
which is Eq.~(\ref{eq:eiv-prop}) of the main text.

\subsection{From the expected loss to the augmented Hessian}

Each squared Frobenius norm in Eq.~(\ref{eq:cmpq-obj}) can be written as a
quadratic form in $\delta W$ via the identity $\|A B\|_F^2 = \mathrm{tr}(A B B^\top A^\top)$:
\begin{align}
\|\delta W\,X\|_F^2 &\;=\; \mathrm{tr}\bigl(\delta W\,(X X^\top)\,\delta W^\top\bigr) \;=\; \mathrm{tr}\bigl(\delta W\,H\,\delta W^\top\bigr), \\
\|\delta W\,\Delta X\|_F^2 &\;=\; \mathrm{tr}\bigl(\delta W\,(\Delta X\,\Delta X^\top)\,\delta W^\top\bigr) \;=\; \mathrm{tr}\bigl(\delta W\,V\,\delta W^\top\bigr).
\end{align}
Adding the two with weight $\lambda$ on the second:
\begin{equation}
\|\delta W\,X\|_F^2 + \lambda\,\|\delta W\,\Delta X\|_F^2
\;=\; \mathrm{tr}\bigl(\delta W\,(H + \lambda V)\,\delta W^\top\bigr)
\;=\; \mathrm{tr}\bigl(\delta W\,\hat H\,\delta W^\top\bigr),
\end{equation}
which identifies $\hat H = H + \lambda V$ as the curvature of the CMPQ
objective with respect to $\delta W$, as claimed in
Eq.~(\ref{eq:aug-hessian}).

\subsection{Equivalence to the widened-calibration form}

The same Hessian arises from the \textbf{augmented calibration matrix}
\begin{equation}
\label{eq:aug-calib}
\tilde X \;\equiv\; \bigl[\,X \;\big|\; \sqrt{\lambda}\,\Delta X\,\bigr]
\;\in\; \mathbb{R}^{d_{\mathrm{in}} \times 2N}.
\end{equation}
By block multiplication,
\begin{equation}
\tilde X\,\tilde X^\top
\;=\; X X^\top + (\sqrt{\lambda}\,\Delta X)(\sqrt{\lambda}\,\Delta X)^\top
\;=\; X X^\top + \lambda\,\Delta X\,\Delta X^\top
\;=\; H + \lambda V \;=\; \hat H,
\end{equation}
and similarly $\|\delta W\,\tilde X\|_F^2 = \|\delta W\,X\|_F^2 + \lambda\,\|\delta W\,\Delta X\|_F^2$.
Geometrically, $\tilde X$ stacks the $N$ real calibration tokens with $N$
virtual noise samples scaled by $\sqrt{\lambda}$, a finite-sample realization
of the EIV perturbation prior in Eq.~(\ref{eq:eiv-perturb}).
Running vanilla GPTQ on $\tilde X$ is therefore equivalent to running CMPQ
on $X$ with damping $\lambda$, requiring no modification to the solver.

\subsection{Connection to anisotropic Tikhonov regularization}

Viewed as a regularizer, $\hat H = H + \lambda V$ instantiates
\emph{generalized Tikhonov} on the GPTQ Hessian with regularization operator
$\Delta X^\top$, in contrast to ordinary ridge ($V = I$) which is agnostic to
where upstream noise lies in input space. GPTQ itself uses ordinary ridge in
the form of a small numerical floor $\epsilon I$ added to $H$ for Cholesky
stability; CMPQ replaces this isotropic floor with a direction-aware operator
that concentrates penalty along the empirical noise subspace spanned by
$\Delta X$.





%% file: sections/appendix-additional.tex
\section{Additional Experimental Results}
\label{app:additional}

All runs in this appendix use the protocol of
Section~\ref{sec:exp-setup} (A100 GPUs, $128 \times 2048$ C4
calibration sequences, group size $128$, no BOS token).

\subsection{Calibration-Seed Variability}
\label{app:seeds}

Table~\ref{tab:seeds} repeats the 2-bit comparison between GPTAQ and
CMPQ ($\lambda{=}0.01$, the WikiText-2 choice for both models) over
four calibration seeds (42, 0, 1, 7); seed 42 is the run reported in
Table~\ref{tab:main-w2}.

\begin{table}[h]
\centering
\footnotesize
\begin{tabular}{ll ccc}
\toprule
Model (2-bit) & Method & WT2 $\downarrow$ & C4 $\downarrow$ & Avg $\uparrow$ \\
\midrule
\multirow{2}{*}{Qwen3-8B}
  & GPTAQ & $27.54 \pm 4.28$ & $24.43 \pm 1.06$ & $52.59 \pm 1.65$ \\
  & CMPQ  & $\mathbf{22.47 \pm 0.88}$ & $\mathbf{22.00 \pm 0.49}$ & $\mathbf{55.78 \pm 4.43}$ \\
\midrule
\multirow{2}{*}{Qwen3-14B}
  & GPTAQ & $\mathbf{14.75 \pm 0.45}$ & $\mathbf{16.26 \pm 0.23}$ & $62.82 \pm 0.54$ \\
  & CMPQ  & $17.23 \pm 3.14$ & $17.28 \pm 0.38$ & $\mathbf{64.89 \pm 1.22}$ \\
\bottomrule
\end{tabular}
\caption{2-bit results over four calibration seeds (mean $\pm$ sample
standard deviation, $n{=}4$).}
\label{tab:seeds}
\end{table}

\subsection{Damping Structure}
\label{app:damping}

Table~\ref{tab:damping} compares three damping matrices $D$ of equal
trace, $\mathrm{tr}(D) = \lambda\,\mathrm{tr}(H)$, added to the GPTQ
Hessian on Qwen3-8B at 2-bit: isotropic
($\lambda\,\mathrm{tr}(H)/d\cdot I$), diagonal
($\lambda\,r\,\mathrm{diag}(V)$ with $r = \mathrm{tr}(H)/\mathrm{tr}(V)$), and
the full CMPQ term $\lambda\,r\,V$. All other settings are
identical, including the MSE clip search.

\begin{table*}[h]
\centering
\footnotesize
\begin{tabular}{ll cc cccccc c}
\toprule
& & \multicolumn{2}{c}{PPL $\downarrow$} & \multicolumn{6}{c}{Zero-shot Acc.\ (\%) $\uparrow$} & \\
\cmidrule(lr){3-4} \cmidrule(lr){5-10}
$\lambda$ & Damping & WT2 & C4 & ARC-c & ARC-e & BoolQ & HS & PIQA & WG & Avg$\uparrow$ \\
\midrule
\multirow{3}{*}{$0.01$}
  & Isotropic & 24.38 & 22.85 & \textbf{37.46} & \textbf{63.72} & \textbf{69.48} & 52.94 & 69.15 & \textbf{61.09} & \textbf{58.97} \\
  & Diagonal  & 23.49 & 22.65 & 34.73 & 61.41 & 65.29 & 53.16 & 67.36 & 61.56 & 57.25 \\
  & Full $V$  & \textbf{23.01} & \textbf{22.37} & 36.60 & 63.09 & 65.69 & \textbf{53.50} & \textbf{69.86} & 59.75 & 58.08 \\
\midrule
\multirow{3}{*}{$0.1$}
  & Isotropic & 31.62 & 27.69 & 29.52 & 45.03 & 64.25 & 47.50 & 63.49 & 56.35 & 51.03 \\
  & Diagonal  & 27.85 & 26.02 & 35.32 & 61.99 & 64.71 & 49.99 & 69.42 & 58.33 & 56.63 \\
  & Full $V$  & \textbf{23.74} & \textbf{22.99} & \textbf{38.57} & \textbf{65.45} & \textbf{66.85} & \textbf{54.25} & \textbf{69.91} & \textbf{61.17} & \textbf{59.37} \\
\midrule
\multirow{3}{*}{$1$}
  & Isotropic & 49.48 & 45.05 & 29.27 & 48.44 & 62.69 & 41.48 & 63.49 & 51.70 & 49.51 \\
  & Diagonal  & 47.03 & 39.92 & 31.40 & 55.81 & 64.31 & 42.29 & 65.45 & 52.01 & 51.88 \\
  & Full $V$  & \textbf{27.10} & \textbf{26.25} & \textbf{39.42} & \textbf{65.07} & \textbf{67.19} & \textbf{53.84} & \textbf{70.29} & \textbf{60.06} & \textbf{59.31} \\
\bottomrule
\end{tabular}
\caption{Damping-structure ablation on Qwen3-8B at 2-bit with matched
trace. Best per column within each $\lambda$ block is in \textbf{bold}.}
\label{tab:damping}
\end{table*}

\subsection{Other Model Families}
\label{app:families}

Table~\ref{tab:other-families} applies the same protocol to
Qwen2.5-7B and Mistral-Nemo-Base-2407 (12B) at 2-bit, with $\lambda$
chosen by WikiText-2 perplexity from $\{0.01, 0.1, 0.5, 1\}$. On
Qwen2.5-7B, CMPQ improves over GPTAQ on WikiText-2 perplexity and on
the zero-shot average. On Mistral-Nemo, CMPQ improves the zero-shot
average, while GPTAQ has the lower perplexity.

\begin{table*}[h]
\centering
\footnotesize
\setlength{\tabcolsep}{4pt}
\begin{tabular}{ll cc cccccc c}
\toprule
& & \multicolumn{2}{c}{PPL $\downarrow$} & \multicolumn{6}{c}{Zero-shot Acc.\ (\%) $\uparrow$} & \\
\cmidrule(lr){3-4} \cmidrule(lr){5-10}
Model (2-bit) & Method & WT2 & C4 & ARC-c & ARC-e & BoolQ & HS & PIQA & WG & Avg$\uparrow$ \\
\midrule
\multirow{4}{*}{Qwen2.5-7B}
  & BF16  & 6.85 & 10.39 & 51.19 & 77.23 & 84.71 & 78.85 & 79.82 & 72.61 & 74.07 \\
  & GPTQ  & 64.47 & 53.07 & 24.49 & 29.63 & 46.39 & 36.65 & 55.82 & 49.57 & 40.42 \\
  & GPTAQ & 28.55 & \textbf{21.72} & 30.29 & 57.03 & 63.98 & 53.12 & 68.34 & 60.22 & 55.49 \\
  & CMPQ ($\lambda{=}0.1$) & \textbf{19.99} & 22.28 & \textbf{33.11} & \textbf{59.85} & \textbf{67.03} & \textbf{57.69} & \textbf{70.35} & \textbf{61.17} & \textbf{58.20} \\
\midrule
\multirow{4}{*}{Mistral-Nemo-12B}
  & BF16  & 5.75 & 8.41 & 59.47 & 83.33 & 85.87 & 83.06 & 82.32 & 76.40 & 78.41 \\
  & GPTQ  & 5741 & 317.2 & 22.70 & 28.49 & 41.47 & 28.22 & 51.85 & 48.93 & 36.94 \\
  & GPTAQ & \textbf{42.41} & \textbf{24.61} & 24.66 & 43.10 & 66.15 & \textbf{51.79} & \textbf{64.31} & 55.25 & 50.88 \\
  & CMPQ ($\lambda{=}0.01$) & 66.93 & 40.79 & \textbf{28.75} & \textbf{50.13} & \textbf{67.58} & 49.67 & 63.22 & \textbf{58.96} & \textbf{53.05} \\
\bottomrule
\end{tabular}
\caption{2-bit results on other model families. Best quantized result
per column within each model is in \textbf{bold}.}
\label{tab:other-families}
\end{table*}

\subsection{Calibration Cost}
\label{app:efficiency}

Table~\ref{tab:wallclock} lists the measurements summarized in
Section~\ref{sec:exp-complexity}. Wall-clock time was measured on
shared nodes and varies by about $10$--$20\%$ between runs of the same
configuration; peak allocated memory is identical across runs.

\begin{table}[h]
\centering
\footnotesize
\begin{tabular}{l cc cc}
\toprule
& \multicolumn{2}{c}{Qwen3-8B} & \multicolumn{2}{c}{Qwen3-14B} \\
\cmidrule(lr){2-3} \cmidrule(lr){4-5}
Method & Time (min) & Mem.\ (GB) & Time (min) & Mem.\ (GB) \\
\midrule
GPTQ  & $47.3$ & $9.2$ & $79.7$ & $13.6$ \\
GPTAQ & $39.7 \pm 5.0$ & $9.8$ & $69.1 \pm 3.6$ & $14.8$ \\
CMPQ  & $46.5 \pm 6.0$ & $10.3$ & $78.8 \pm 9.6$ & $15.9$ \\
\bottomrule
\end{tabular}
\caption{Quantization wall-clock time and peak allocated GPU memory
at 2-bit on a single A100 (mean $\pm$ std over $3$ calibration seeds; GPTQ is a single run).}
\label{tab:wallclock}
\end{table}

\clearpage

%% file: reference.bib
@article{dettmers2022gpt3,
  title={{LLM.int8()}: 8-bit matrix multiplication for transformers at scale},
  author={Dettmers, Tim and Lewis, Mike and Belkada, Younes and Zettlemoyer, Luke},
  journal={Advances in neural information processing systems},
  volume={35},
  pages={30318--30332},
  year={2022},
  eprint={2208.07339},
  archivePrefix={arXiv},
  url={https://proceedings.neurips.cc/paper_files/paper/2022/hash/c3ba4962c05c49636d4c6206a97e9c8a-Abstract-Conference.html}
}

@article{frantar2022gptq,
  title={Gptq: Accurate post-training quantization for generative pre-trained transformers},
  author={Frantar, Elias and Ashkboos, Saleh and Hoefler, Torsten and Alistarh, Dan},
  journal={arXiv preprint arXiv:2210.17323},
  year={2022}
}

@article{frantar2022optimal,
  title={Optimal brain compression: A framework for accurate post-training quantization and pruning},
  author={Frantar, Elias and Alistarh, Dan},
  journal={Advances in Neural Information Processing Systems},
  volume={35},
  pages={4475--4488},
  year={2022}
}

@article{lin2024awq,
  title={Awq: Activation-aware weight quantization for on-device llm compression and acceleration},
  author={Lin, Ji and Tang, Jiaming and Tang, Haotian and Yang, Shang and Chen, Wei-Ming and Wang, Wei-Chen and Xiao, Guangxuan and Dang, Xingyu and Gan, Chuang and Han, Song},
  journal={Proceedings of machine learning and systems},
  volume={6},
  pages={87--100},
  year={2024}
}

@article{dettmers2023spqr,
  title={Spqr: A sparse-quantized representation for near-lossless llm weight compression},
  author={Dettmers, Tim and Svirschevski, Ruslan and Egiazarian, Vage and Kuznedelev, Denis and Frantar, Elias and Ashkboos, Saleh and Borzunov, Alexander and Hoefler, Torsten and Alistarh, Dan},
  journal={International Conference on Learning Representations},
  year={2024},
  eprint={2306.03078},
  archivePrefix={arXiv},
  url={https://openreview.net/forum?id=Q1u25ahSuy}
}

@inproceedings{li2021brecq,
  title={{BRECQ}: Pushing the limit of post-training quantization by block reconstruction},
  author={Li, Yuhang and Gong, Ruihao and Tan, Xu and Yang, Yang and Hu, Peng and Zhang, Qi and Yu, Fengwei and Wang, Wei and Gu, Shi},
  booktitle={International Conference on Learning Representations},
  year={2021},
  eprint={2102.05426},
  archivePrefix={arXiv},
  url={https://openreview.net/forum?id=POWv6hDd9XH}
}

@inproceedings{nagel2020up,
  title={Up or down? adaptive rounding for post-training quantization},
  author={Nagel, Markus and Amjad, Rana Ali and Van Baalen, Mart and Louizos, Christos and Blankevoort, Tijmen},
  booktitle={International conference on machine learning},
  pages={7197--7206},
  year={2020},
  organization={PMLR}
}

@inproceedings{tseng2024quip,
  title={{QuIP\#}: Even better {LLM} quantization with Hadamard incoherence and lattice codebooks},
  author={Tseng, Albert and Chee, Jerry and Sun, Qingyao and Kuleshov, Volodymyr and De Sa, Christopher},
  booktitle={Proceedings of the 41st International Conference on Machine Learning},
  series={Proceedings of Machine Learning Research},
  volume={235},
  pages={48630--48656},
  year={2024},
  publisher={PMLR},
  url={https://proceedings.mlr.press/v235/tseng24a.html}
}

@article{chee2023quip,
  title={Quip: 2-bit quantization of large language models with guarantees},
  author={Chee, Jerry and Cai, Yaohui and Kuleshov, Volodymyr and De Sa, Christopher M},
  journal={Advances in neural information processing systems},
  volume={36},
  pages={4396--4429},
  year={2023}
}

@article{grattafiori2024llama,
  title={The llama 3 herd of models},
  author={Grattafiori, Aaron and Dubey, Abhimanyu and Jauhri, Abhinav and Pandey, Abhinav and Kadian, Abhishek and Al-Dahle, Ahmad and Letman, Aiesha and Mathur, Akhil and Schelten, Alan and Vaughan, Alex and others},
  journal={arXiv preprint arXiv:2407.21783},
  year={2024}
}

@article{guo2025deepseek,
  title={Deepseek-r1: Incentivizing reasoning capability in llms via reinforcement learning},
  author={Guo, Daya and Yang, Dejian and Zhang, Haowei and Song, Junxiao and Wang, Peiyi and Zhu, Qihao and Xu, Runxin and Zhang, Ruoyu and Ma, Shirong and Bi, Xiao and others},
  journal={arXiv preprint arXiv:2501.12948},
  year={2025}
}

@article{yang2025qwen3,
  title={Qwen3 technical report},
  author={Yang, An and Li, Anfeng and Yang, Baosong and Zhang, Beichen and Hui, Binyuan and Zheng, Bo and Yu, Bowen and Gao, Chang and Huang, Chengen and Lv, Chenxu and others},
  journal={arXiv preprint arXiv:2505.09388},
  year={2025}
}

@incollection{gholami2022survey,
  title={A survey of quantization methods for efficient neural network inference},
  author={Gholami, Amir and Kim, Sehoon and Dong, Zhen and Yao, Zhewei and Mahoney, Michael W and Keutzer, Kurt},
  booktitle={Low-power computer vision},
  pages={291--326},
  year={2022},
  publisher={Chapman and Hall/CRC}
}

@inproceedings{xiao2023smoothquant,
  title={Smoothquant: Accurate and efficient post-training quantization for large language models},
  author={Xiao, Guangxuan and Lin, Ji and Seznec, Mickael and Wu, Hao and Demouth, Julien and Han, Song},
  booktitle={International conference on machine learning},
  pages={38087--38099},
  year={2023},
  organization={PMLR}
}

@article{shao2023omniquant,
  title={Omniquant: Omnidirectionally calibrated quantization for large language models},
  author={Shao, Wenqi and Chen, Mengzhao and Zhang, Zhaoyang and Xu, Peng and Zhao, Lirui and Li, Zhiqian and Zhang, Kaipeng and Gao, Peng and Qiao, Yu and Luo, Ping},
  journal={arXiv preprint arXiv:2308.13137},
  year={2023}
}

@article{li2025gptaq,
  title={Gptaq: Efficient finetuning-free quantization for asymmetric calibration},
  author={Li, Yuhang and Yin, Ruokai and Lee, Donghyun and Xiao, Shiting and Panda, Priyadarshini},
  journal={arXiv preprint arXiv:2504.02692},
  year={2025}
}

@article{li2026rethinking,
  title={Rethinking Residual Errors in Compensation-based LLM Quantization},
  author={Li, Shuaiting and Deng, Juncan and Xu, Kedong and Deng, Rongtao and Gu, Hong and Jiang, Minghan and Shen, Haibin and Huang, Kejie},
  journal={arXiv preprint arXiv:2604.07955},
  year={2026}
}

@article{wolf2019huggingface,
  title={Huggingface's transformers: State-of-the-art natural language processing},
  author={Wolf, Thomas and Debut, Lysandre and Sanh, Victor and Chaumond, Julien and Delangue, Clement and Moi, Anthony and Cistac, Pierric and Rault, Tim and Louf, R{\'e}mi and Funtowicz, Morgan and others},
  journal={arXiv preprint arXiv:1910.03771},
  year={2019}
}

@article{paszke2019pytorch,
  title={Pytorch: An imperative style, high-performance deep learning library},
  author={Paszke, Adam and Gross, Sam and Massa, Francisco and Lerer, Adam and Bradbury, James and Chanan, Gregory and Killeen, Trevor and Lin, Zeming and Gimelshein, Natalia and Antiga, Luca and others},
  journal={Advances in neural information processing systems},
  volume={32},
  year={2019}
}

@article{raffel2020exploring,
  title={Exploring the limits of transfer learning with a unified text-to-text transformer},
  author={Raffel, Colin and Shazeer, Noam and Roberts, Adam and Lee, Katherine and Narang, Sharan and Matena, Michael and Zhou, Yanqi and Li, Wei and Liu, Peter J},
  journal={Journal of machine learning research},
  volume={21},
  number={140},
  pages={1--67},
  year={2020}
}

@article{merity2016pointer,
  title={Pointer sentinel mixture models},
  author={Merity, Stephen and Xiong, Caiming and Bradbury, James and Socher, Richard},
  journal={arXiv preprint arXiv:1609.07843},
  year={2016}
}

@article{gao2021framework,
  title={A framework for few-shot language model evaluation},
  author={Gao, Leo and Tow, Jonathan and Biderman, Stella and Black, Sid and DiPofi, Anthony and Foster, Charles and Golding, Laurence and Hsu, Jeffrey and McDonell, Kyle and Muennighoff, Niklas and others},
  journal={Zenodo},
  year={2021}
}
